\documentclass{article}

\usepackage[final]{exait_2025}

\usepackage[utf8]{inputenc} 
\usepackage[T1]{fontenc}    
\usepackage{hyperref}       
\usepackage{url}            
\usepackage{booktabs}       
\usepackage{amsfonts}       
\usepackage{nicefrac}       
\usepackage{microtype}      
\usepackage{xcolor}         
\usepackage{wrapfig}

\usepackage{amsmath}
\usepackage{amssymb}
\usepackage{mathtools}
\usepackage{graphicx}
\usepackage{booktabs}
\usepackage{times}
\usepackage{latexsym}

\usepackage{enumerate}
\usepackage[inline]{enumitem}
\usepackage{amsfonts,dsfont}
\usepackage{nicefrac}
\usepackage{microtype}
\usepackage{graphics}
\usepackage{caption}
\usepackage{subcaption}
\usepackage{wrapfig}
\usepackage{enumitem}
\usepackage[normalem]{ulem}
\usepackage{amssymb}
\usepackage{multicol}
\usepackage{adjustbox}
\usepackage{booktabs}
\usepackage{multirow}
\usepackage{array}
\usepackage{color}
\usepackage{xspace}
\usepackage[capitalize,noabbrev]{cleveref}

\newcommand{\pjn}{{Wild-P-Diff}}

\title{Reimagining Parameter Space Exploration with Diffusion Models}

\author{%
  Lijun Zhang\\
  University of Massachusetts Amherst\\
  \texttt{lijunzhang@cs.umass.edu} \\
  \And
  Xiao Liu\\
  University of Massachusetts Amherst\\
  \texttt{xiaoliu1990@umass.edu} \\
  \And
  Hui Guan\\
  University of Massachusetts Amherst\\
  \texttt{huiguan@cs.umass.edu} \\
}

\begin{document}

\maketitle

\begin{abstract}
Adapting neural networks to new tasks typically requires task-specific fine-tuning, which is time-consuming and reliant on labeled data. We explore a generative alternative that produces task-specific parameters directly from task identity, eliminating the need for task-specific training.
To this end, we propose using diffusion models to learn the underlying structure of effective task-specific parameter space and synthesize parameters on demand. 
Once trained, the task-conditioned diffusion model can generate specialized weights directly from task identifiers.
We evaluate this approach across three scenarios: generating parameters for a single seen task, for multiple seen tasks, and for entirely unseen tasks. 
Experiments show that diffusion models can generate accurate task-specific parameters and support multi-task interpolation when parameter subspaces are well-structured, but fail to generalize to unseen tasks, highlighting both the potential and limitations of this generative solution.
\end{abstract}

\section{Introduction}\label{sect:pdiff-intro}
Exploring high-dimensional parameter spaces is a fundamental problem in machine learning~\cite{bengio2009learning}.
A core challenge is to identify a set of parameters that enable high task performance -- that is, parameters that encode inductive biases aligned with the target task and generalize well to new inputs. 
The traditional approach for this parameter exploration is gradient-based optimization, such as stochastic gradient descent (SGD), which iteratively updates parameters to minimize task-specific loss. 
For example, when adapting a pre-trained network to a new task, it is common to fine-tune lightweight, task-specific modules (e.g., adapters or heads) via gradient descent on task-specific data~\cite{hu2022lora}.

While effective, this approach has two key limitations. 
First, fine-tuning is time-consuming, often requiring multiple training epochs per task. Second, it relies heavily on access to sufficient labeled training data for each new task, an assumption that may not hold in practice, especially in low-resource or privacy-sensitive settings.

In this work, we explore a \textbf{generative alternative} to task-specific fine-tuning. 
Instead of using gradient descent to search for parameters that perform well on a given task, we ask: \textbf{can we learn a generative model that directly outputs good parameters for a task, conditioned only on a task description?} 
We hypothesize that diffusion models, a class of generative models known for learning complex, high-dimensional distributions, can capture the underlying structure of effective task-specific parameter spaces. 
If this hypothesis holds, then we can bypass optimization entirely: given only a task identifier or embedding, we can sample well-performed parameters directly, enabling efficient task adaptation without access to task-specific training data.

To test this hypothesis, we frame our investigation around three key research questions:
\begin{itemize}[noitemsep,nolistsep,leftmargin=*,topsep=0pt]
\item \textbf{RQ1 (Task-Specific Generation)}: Can diffusion models generate accurate parameters when conditioned on known tasks?
\item \textbf{RQ2 (Inter-Task Interpolation)}: Can they generate parameters that generalize across multiple tasks by blending task conditions?
\item \textbf{RQ3 (Unseen Task Generalization)}: Can they generate parameters for entirely new, unseen tasks?
\end{itemize}

We study this problem in the context of wildlife classification from camera trap data~\cite{rastikerdar2024situ}, a real-world scenario that naturally induces task-specific variation. 
Camera traps, widely used in ecological studies, are deployed in diverse environments, each with its own background, lighting, and animal distributions, effectively making each location a distinct domain or task. 
This domain is well-suited for evaluating task-specific adaptation: it features inter-task variability, limited labeled data per task, and a strong need for efficient, on-device adaptation.
In practice, parameter-efficient fine-tuning (PEFT) methods such as LoRA are commonly used~\cite{hu2022lora}, and our study explores the possibility of replacing this process with generative parameter inference.

Our experimental results yield three key findings:
\begin{itemize}[noitemsep,nolistsep,leftmargin=*,topsep=0pt]
    \item \textbf{Finding 1 (RQ1)}: Diffusion models can generate task-specific adapter parameters that achieve high accuracy on individual tasks seen during training.
    \item \textbf{Finding 2 (RQ2)}: Interpolating across task conditions enables generating parameters that generalize across multiple related tasks, especially when the underlying parameter subspaces are aligned.
    \item \textbf{Finding 3 (RQ3)}: Generalization to unseen tasks remains limited, highlighting challenges in modeling out-of-distribution task embeddings.
\end{itemize}

These findings suggest that diffusion-based parameter generation is a promising direction for scalable, data-free task adaptation. 
While generalization to unseen tasks beyond the training distribution remains an open challenge, our findings demonstrate the feasibility of using generative models as a new tool for efficient parameter space exploration.
\section{Preliminary and Related Work}\label{sect:background}
We begin by defining the parameter space exploration problem for task specialization, followed by a review of relevant research on parameter generation.

\textbf{Task-Specific Parameter Space Exploration.} 
Parameter space exploration aims to identify task-specific parameters that yield strong performance on a given task. In this setting, the \textbf{search space} is the high-dimensional space of all possible task-specific parameter vectors.
Formally, given a model architecture $f$, task-specific data $\mathcal{D}_{\tau}$, and a loss function $\mathcal{L}_{\tau}$, our goal is to identify parameters $\theta_\tau$ that minimize the task loss,
$\min_{\theta_\tau} \mathcal{L}_{\tau}(f_{\theta_\tau}, \mathcal{D}_{\tau})$,
where $\theta_\tau$ denotes task-adaptive parameters that specialize the shared model $f$ to task $\tau$. 
These task-specific parameters typically correspond to lightweight modules such as prediction heads or parameter-efficient adapters.

\textbf{Parameter Generation.} 
Generating model parameters has been explored in various contexts. 
Early approaches like HyperNetworks~\cite{ha2016hypernetworks} learn to generate weights dynamically for variable architectures. 
SMASH \cite{brock2017smash} introduces a memory-based generation scheme for architecture search.
More recently, G.pt~\cite{peebles2022learning} employs diffusion models to generate new weights conditioned on existing parameters and target objectives.
However, their generated models often underperform compared to directly trained counterparts.
Recent studies~\cite{wang2024neural, wang2025sindiffusion, soro2025diffusion} have pioneered the use of diffusion models for high-performance parameter generation, showing that generated parameters can achieve accuracy comparable to explicitly trained ones.
Unlike these works that emphasize generative quality, our work examines diffusion models' ability to generate parameters under interpolated and unseen task conditions, which holds the potential to become data- and optimization-free alternative to fine-tuning.



\section{Explored Method: \pjn{}}
This section introduces \pjn{} (Wildlife Classification Parameter Diffusion Model), our proposed framework for task-specific parameter generation. Our design follows the structure of latent diffusion models commonly used for image synthesis~\cite{rombach2022high}, but repurposes them to treat model parameters as a new generative modality.

\subsection{Overview of \pjn{}}
\begin{figure}[t]
    \centering
    \includegraphics[width=0.75\textwidth]{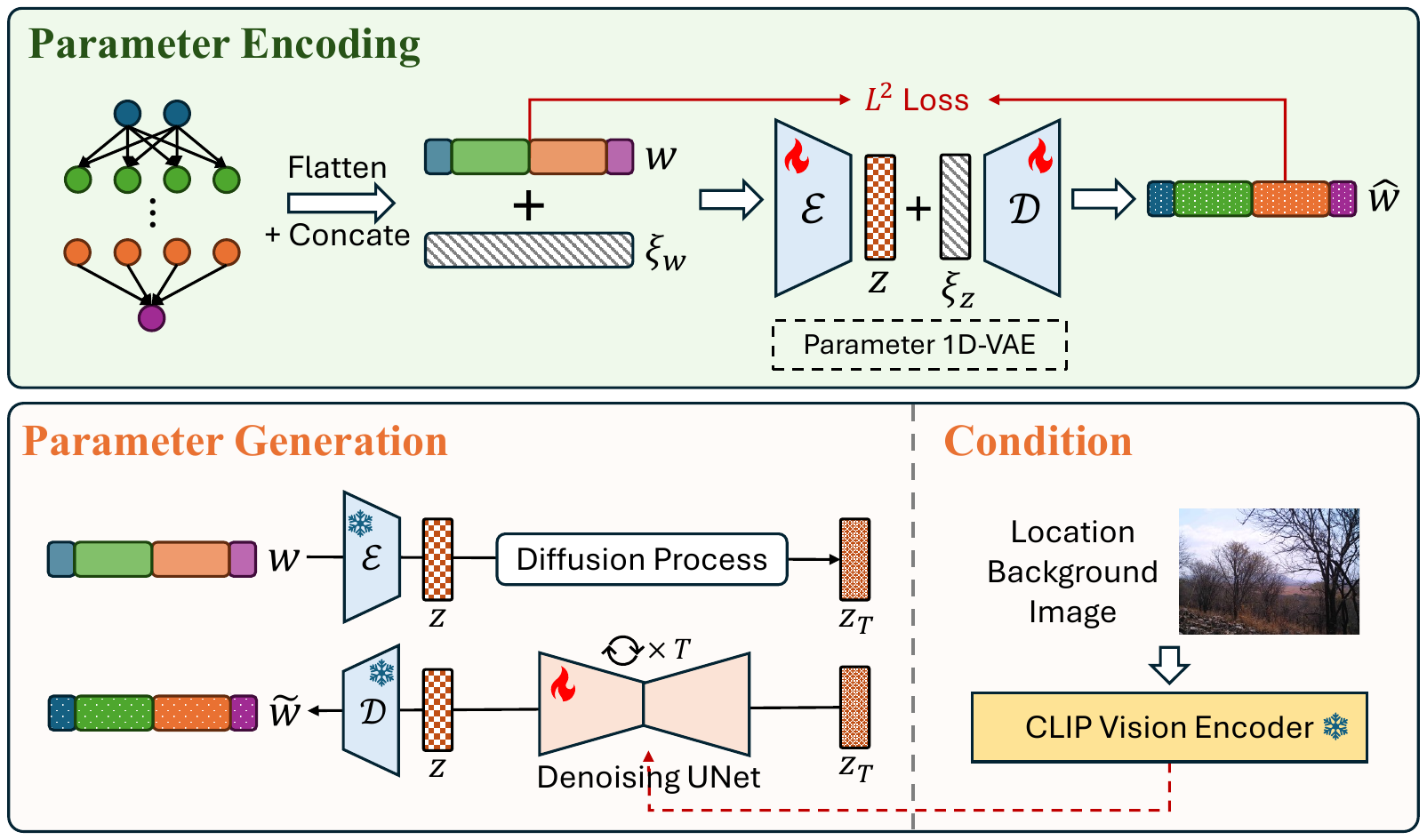}
    \caption{The framework of \pjn{} with two processes: parameter encoding and parameter generation. A parameter VAE aims to extract the latent representations and reconstruct model parameters via the decoder. The extracted representations are used to train a denoising UNet of the diffusion model. The conditional version aims to synthesize high-performance parameters based on the CLIP representation of a specific location background image.
    }
    \vspace{-15pt}
    \label{fig:pdiff-overview}
\end{figure}

Figure~\ref{fig:pdiff-overview} presents the overall structure of the \pjn{} framework, which consists of two main components: \textbf{Parameter Encoding} and \textbf{Parameter Generation}, along with an optional conditioning mechanism for task-specific adaptation.

The parameter encoding process begins with a set of high-performing task-specific parameters. These parameters are flattened and concatenated across layers into a one-dimensional vector, which is then passed through a parameter encoder to obtain a compact latent representation. A corresponding decoder is trained to reconstruct the original parameters from this latent space.

To enable parameter generation, a diffusion model is trained in the latent space to synthesize embeddings of effective task-specific parameters. 
To support conditional generation, task-related information is incorporated into the denoising UNet of the diffusion model during training. 
At inference time, the model samples a latent vector conditioned on task context, which is then decoded into a parameter vector using the trained decoder.

As introduced in Section~\ref{sect:pdiff-intro}, we evaluate our approach in the context of wildlife classification from camera trap data, in which tasks correspond to different deployment locations. 
Hence, the goal becomes to generate location-specific parameters conditioned on location's visual context. Specifically, we use a pre-trained CLIP vision encoder to extract features from the background image of the camera trap site. These embeddings are then injected into the denoising UNet, guiding the generation toward location-adaptive parameters.

\subsection{Parameter Encoding}
To enable parameter generation in the latent space, we first train a parameter Variational Autoencoder (VAE) that maps between full parameter vectors and their latent representations.

\textbf{Training Dataset.} 
To construct the training dataset, we fine-tune the selected location-specific parameters of a pre-trained wildlife classification model across different locations. 
For each of the $T$ locations, we save $N$ fine-tuned checkpoints, yielding a dataset of parameter vectors denoted as $\Theta=[\theta_1,...,\theta_n,...,\theta_{N\times T}]$.
Details of the task-specific parameters collection procedure are provided in Sections~\ref{sect:pdiff-exp-setting}.

Each parameter vector $\theta_n$ is flattened and concatenated across layers into a single one-dimensional vector $w_n\in \mathbb{R}^{K\times1}$, where $K$ is the total number of parameters.
To standardize the data, we apply Z-score normalization~\cite{fei2021z} independently to each layer following~\cite{wang2024neural}.

\textbf{Encoding Architecture.}
The VAE is trained from scratch to encode the parameter vector into a latent space and reconstruct it from its latent representation. The encoding and decoding processes are defined as:
\begin{equation}\label{eq:paramVAE}
    \begin{aligned}
        &z_n = \mathcal{E}(w_n + \xi_{w_n}), \quad \xi_{w_n} \sim \mathcal{N}(0, \sigma^2_w\mathbf{I}); 
        &\hat{w}_n = \mathcal{D}(z_n + \xi{z_n}), \quad \xi_{z_n} \sim \mathcal{N}(0, \sigma^2_z\mathbf{I}),
    \end{aligned}
\end{equation}
where $\mathcal{E}$ and $\mathcal{D}$ represent the encoder and decoder respectively, and $\hat{w}_n$ denotes the reconstructed output from the decoder.
To enhance the robustness and generalization capabilities of the VAE, Gaussian noise is added to both the input vector $w_n$ and the latent representation $z_n$ during training, with noise levels 
controlled by $\sigma^2_w$ and $\sigma^2_z$. The VAE is optimized using an $L^2$ reconstruction loss between $w_n$ and $\hat{w}_n$.

\subsection{Parameter Generation}\label{sect:param-gen}
To enable task-specific parameter synthesis, we train a diffusion model to generate latent parameter representations following the denoising diffusion implicit model (DDIM) \cite{song2020denoising}. Background on diffusion models is provided in Appendix Section~\ref{app:diffusion}.

During training, the diffusion process progressively adds noise to the latent representation $\mathbf{Z}$ obtained from the parameter encoder, transforming it into a noisy latent state $\mathbf{Z}_T$. 
The denoising UNet then learns to recover the clean latent representation $\mathbf{Z}$ from $\mathbf{Z}_T$ through an iterative refinement process spanning $T$ timesteps. 
Once trained, the model can generate new latent by sampling from a Gaussian prior and reversing the diffusion process. The generated latent is then decoded into a parameter vector using the trained parameter decoder.
These generated task-specific parameters are combined with the fixed, task-agnostic backbone of the wildlife classification model to produce a complete model for downstream evaluation.

The core component of the diffusion model is the denoising UNet, specifically adapted for processing the latent of neural network parameters. 
Since parameter vectors lack inherent spatial structure~\cite{wang2024neural}, we replace conventional 2D convolutions in diffusion models for image generation with 1D convolutional blocks throughout the U-Net architecture.
For location-aware parameter generation, we incorporate a conditioning mechanism using contextual visual cues. Specifically, a pre-trained and frozen CLIP vision encoder extracts semantic features from the background image corresponding to each camera trap deployment site. 
These features are injected into the denoising U-Net by directly adding them to the input latent $\mathbf{Z}_t$, following the same positional encoding strategy used for timestep conditioning. 

\looseness-1
\section{Experimental Results}\label{sect:pdiff-exp}
We first validate the capability of \pjn{} to generate high-performing parameters without conditioning, assessing whether it can model and reproduce the underlying parameter distribution (Section~\ref{sect:exp-uncond-param}). 
We then address the three research questions posed in Section~\ref{sect:pdiff-intro} (Section~\ref{sect:exp-cond-param}).

\subsection{Experiments Settings}\label{sect:pdiff-exp-setting}

\textbf{Wildlife Classification Dataset and Architecture.}
We conduct our experiments on the \textbf{Serengeti Safari Camera Trap} dataset~\cite{swanson2015snapshot}, a benchmark for wildlife monitoring. 
This dataset contains wildlife images, including an “empty” class representing background scenes without animals. 
Following~\cite{rastikerdar2024situ}, we select the 18 most frequent animals along with the empty class, forming a \textbf{19-way classification task}.
As summarized in Appendix Table~\ref{tab:dataset_stats}, the dataset is split by location. 
The \textit{train} set includes 110 locations and is used to train the wildlife classification model and construct the parameter dataset for the VAE and diffusion UNet in \pjn{}. 
The \textit{test} set includes 5 locations, which are used to assess generalization to unseen tasks.
We adopt \textbf{EfficientNet-B0}~\cite{tan2019efficientnet} pre-trained on ImageNet as the backbone for the wildlife classification model.

\textbf{Parameter Dataset Preparation.}
To train the parameter VAE and the diffusion model in \pjn{}, we construct a dataset of task-specific parameters across locations.
We begin by pretraining the classification model on images from the first 100 locations in the train set. Since the dataset is imbalanced, with most images belonging to the empty class, we apply an upsampling strategy by reweighting samples based on class frequency during training.

We then fine-tune only the \textbf{LoRA adapters of the first six layers} on the remaining 10 locations to obtain location-specific parameters. All other weights in EfficientNet are frozen. 
Table~\ref{tab:finetune-locs} in the Appendix provides detailed statistics for these locations, where 10\% of each location's data is reserved for evaluation on seen tasks.  Justification for the selected parameter subset is provided in Section~\ref{sect:what-param}.

To form the training dataset for \pjn{}, we continue fine-tuning beyond convergence and save 300 checkpoints per location, yielding a total of 3,000 parameter vectors. 
In practice, we set the interval at which fine-tuned checkpoints are saved, denoted as the \emph{Saving Interval}, as 100 unless noted differently to maintain high intra-location diversity for the saved checkpoints (see analysis in Section~\ref{sect:exp-uncond-param}).
For the conditional generation setting, we assign each location a representative background image to serve as the conditioning input, capturing the environmental context of the deployment site.

\textbf{Training and Inference.}
The detailed architectures for the parameter VAE and diffusion UNet are provided in Appendix~\ref{app:architecture}, and key training hyperparameters are summarized in Table~\ref{tab:training_recipes}.
Besides, for the VAE, the noise scales $\sigma_w$ and $\sigma_z$ in Equation~\ref{eq:paramVAE} are set to 0.001 and 0.1, respectively, to control the Gaussian noise added to the input and latent representations during training.

During inference, we generate 100 sets of task-specific parameters by sampling random noise inputs and passing them through the trained diffusion model and decoder. 
For conditional generation, location-specific embeddings are injected into the diffusion model, guiding the synthesis of location-adapted parameters. 
The generated parameters are then combined with the fixed, location-agnostic backbone weights to form a complete classification model.

\textbf{Baselines.}
We compare \pjn{} with:
\begin{itemize}[noitemsep,nolistsep,leftmargin=*,topsep=0pt]
    \item \textbf{Pretrain:} The classification accuracy of the original pre-trained model on each location.
    \item \textbf{FTed:} The average accuracy of all fine-tuned checkpoints for a given location.
    \item \textbf{Ensemble:} Accuracy from the weighted ensemble of the fine-tuned checkpoints~\cite{wortsman2022model}.
\end{itemize}

\textbf{Metrics.}
We report the mean and standard deviation of classification accuracy on each location’s validation set, capturing the overall performance of both fine-tuned and generated models. 
To assess parameter similarity, we use \emph{cosine similarity} as a distance metric. 

\
\subsection{Unconditional Parameter Generation}\label{sect:exp-uncond-param}
\begin{wraptable}{r}{0.5\textwidth}
\vspace{-.4cm}
\centering
\scriptsize
\tabcolsep=0.08cm
\caption{The mean and standard deviation of classification accuracy for 100 sets of generated parameters, comparing them against two baselines, \emph{Pretrain} and \emph{FTed}. The \textit{Saving Interval} refers to how frequently we save the fine-tuned checkpoints.  
} \label{tab:uncond-acc}
\begin{tabular}{c|c|ccc}
\toprule
\begin{tabular}[c]{@{}l@{}}Saving\\Interval\end{tabular} & Pretrain                & FTed            & \pjn{}    & $\Delta$ Acc.  \\
\midrule
1               & \multirow{3}{*}{81.43} & 92.29 $\pm$ 0.0137 & 93.80 $\pm$ 0.0137 & $+0.0151$\\
10              &                         & 92.68 $\pm$ 0.0139 & 93.66 $\pm$ 0.0067 & $+0.0098$ \\
100             &                         & 94.19 $\pm$ 0.0147 & 93.80 $\pm$ 0.0137 & $-0.0039$\\
\bottomrule
\end{tabular}
\vspace{-.3cm}
\end{wraptable}

We begin by evaluating the unconditional version of \pjn{}, trained on task-specific parameters from a single location, \emph{R10}. Since the diffusion model is trained exclusively on parameters fine-tuned for this location, we expect it to generate weights that perform well on the same task. 
Table~\ref{tab:uncond-acc} reports the mean and standard deviation of classification accuracy over 100 generated parameter samples.
We compare task performance against two baselines: \emph{Pretrain}, the original pre-trained model without task adaptation, and \emph{FTed}, the average performance of fine-tuned checkpoints.
Remarkably, the generated parameters match the accuracy of the fine-tuned checkpoints on R10.

To further investigate the model’s generative behavior, we vary the diversity of the training set by evaluating different \emph{Saving Interval}. 
Specifically, a smaller interval (e.g., every 10 iterations) yields many similar checkpoints, while a larger interval (e.g., every 100 iterations) results in more diverse training samples (see Figure~\ref{fig:uncond-input-diversity} in the Appendix). 
\pjn{} consistently generates parameters with competitive accuracy as the fine-tuned checkpoints regardless of the input diversity.
More importantly, as shown in Figure~\ref{fig:output_novelty_diversity} in the Appendix, we find that higher input diversity leads to more \textit{novel} parameter generation in \pjn{}. 
When the training set contains more distinct parameter instances, the diffusion model is better able to generalize beyond memorized patterns and synthesize new parameters different from the training samples.
Detailed analysis of this diversity effect is presented in Appendix~\ref{sect:exp-uncond-param-cont}. 

\subsection{Conditional Parameter Generation}\label{sect:exp-cond-param}
\textbf{Parameters Similarity across Locations.} 
This section answers the three research questions in Section~\ref{sect:pdiff-intro}. 
Specifically, we investigate how the similarity between task-specific parameters from different locations affects the model's ability to generate new parameters under conditioning, as detailed in Appendix~\ref{sect:exp-cond-param-cont}.
\begin{itemize}[noitemsep,nolistsep,leftmargin=*,topsep=0pt]
    \item \textbf{Low Similarity (L)}: Task-specific parameter sets across locations are nearly orthogonal, with cosine similarity $\approx 0$.
    \item \textbf{Medium Similarity (M)}: Parameters exhibit moderate similarity across locations, with cosine similarity around 0.5.
    \item \textbf{High Simiarity (H)}: Parameters are highly similar across locations, with cosine similarity approaching 0.98.
\end{itemize}

\textbf{RQ1: Task-Specific Generation.} 
Table~\ref{tab:cond-acc} reports the classification accuracy for each location for both fine-tuned checkpoints and diffusion-generated parameters across all three cross-location similarity settings: \pjn{}-L, \pjn{}-M, and \pjn{}-H. 
As in the unconditional setting, we find that \textbf{task-conditioned diffusion models can generate high-performing parameters for each location regardless of the parameter similarity in the training dataset}, guided by the background images for each location. 
This confirms that the model can differentiate seen tasks and produce task-specific parameters based on contextual input.

\begin{table}[h]
\vspace{-.4cm}
\centering
\scriptsize
\tabcolsep=0.08cm
\caption{The mean and standard deviation of location-specific accuracy on seen locations for 100 sets of generated parameters, comparing them against two baselines, \emph{Pretrain} and \emph{FTed}, across all three cross-location similarity settings (\pjn{}-L, \pjn{}-M, and \pjn{}-H).} \label{tab:cond-acc}
\begin{tabular}{c|c|ccc|ccc|ccc}
\toprule
\multirow{3}{*}{\begin{tabular}[c]{@{}c@{}}Loc. \\ ID\end{tabular}} & \multicolumn{10}{c}{Accuracy}                                                                                                                                                                                                                                 \\ \cmidrule{2-11} 
                             & \multicolumn{1}{c|}{\multirow{2}{*}{Pretrained}} & \multicolumn{3}{c|}{Low Similarity: random LoRA init}                  & \multicolumn{3}{c|}{Medium Similarity: from same LoRA init}            & \multicolumn{3}{c}{High Similarity: from converged LoRA} \\ \cmidrule{3-11} 
                             & \multicolumn{1}{c|}{}                            & FTed         & Wild-P-Diff  & \multicolumn{1}{c|}{$\Delta$ Acc.} & FTed         & Wild-P-Diff  & \multicolumn{1}{c|}{$\Delta$ Acc.} & FTed            & Wild-P-Diff    & $\Delta$ Acc.   \\ \midrule
R10                          & 0.81                                             & 94.19 $\pm$ 1.47 & 93.83 $\pm$ 1.17 & -0.36                                    & 94.66 $\pm$ 1.11 & 94.32 $\pm$ 1.22 & -0.34                                    & 96.48 $\pm$ 0.96    & 95.51 $\pm$ 2.11   & -0.97                 \\
R12                          & 0.85                                             & 93.63 $\pm$ 1.17 & 93.68 $\pm$ 1.04 & +0.05                                     & 93.86 $\pm$ 0.96 & 93.23 $\pm$ 2.49 & -0.63                                    & 95.88 $\pm$ 1.11    & 95.02 $\pm$ 1.44   & -0.86                 \\
U11                          & 0.45                                             & 98.56 $\pm$ 2.17 & 98.51 $\pm$ 3.27 & -0.05                                    & 97.20 $\pm$ 2.45 & 95.41 $\pm$ 5.82 & -1.79                                    & 98.26 $\pm$ 2.31    & 95.54 $\pm$ 4.69   & -2.72                 \\
Q08                          & 0.57                                             & 87.66 $\pm$ 3.29 & 87.40 $\pm$ 2.92 & -0.26                                    & 86.43 $\pm$ 3.17 & 84.46 $\pm$ 5.77 & -1.97                                    & 86.61 $\pm$ 3.10    & 83.99 $\pm$ 3.39   & -2.62                 \\
M08                          & 0.65                                             & 99.93 $\pm$ 0.57 & 99.80 $\pm$ 0.97 & -0.13                                    & 99.90 $\pm$ 0.81 & 99.80 $\pm$ 0.97 & -0.1                                     & 99.91 $\pm$ 0.64    & 99.80 $\pm$ 0.97   & -0.11                 \\
D09                          & 0.87                                             & 97.48 $\pm$ 2.23 & 97.56 $\pm$ 2.83 & +0.08                                     & 97.87 $\pm$ 2.20 & 97.69 $\pm$ 3.10 & -0.18                                    & 98.31 $\pm$ 2.14    & 98.69 $\pm$ 1.99   & +0.38                  \\
H13                          & 0.78                                             & 92.24 $\pm$ 4.10 & 91.16 $\pm$ 4.45 & -1.08                                    & 92.14 $\pm$ 4.26 & 91.16 $\pm$ 4.01 & -0.98                                    & 92.46 $\pm$ 3.37    & 91.27 $\pm$ 3.26   & -1.19                 \\
S09                          & 1.00                                             & 95.89 $\pm$ 4.40 & 97.26 $\pm$ 3.78 & +1.37                                     & 96.53 $\pm$ 4.30 & 95.06 $\pm$ 5.93 & -1.47                                    & 96.84 $\pm$ 3.82    & 96.80 $\pm$ 4.36   & -0.04                 \\
U13                          & 0.17                                             & 99.90 $\pm$ 0.52 & 99.87 $\pm$ 0.53 & -0.03                                    & 99.89 $\pm$ 0.54 & 99.65 $\pm$ 1.19 & -0.24                                    & 99.84 $\pm$ 0.88    & 99.73 $\pm$ 0.96   & -0.11                 \\
S12                          & 0.95                                             & 98.81 $\pm$ 2.06 & 98.80 $\pm$ 2.27 & -0.01                                    & 99.13 $\pm$ 1.84 & 98.76 $\pm$ 2.39 & -0.37                                    & 97.55 $\pm$ 2.37    & 97.23 $\pm$ 2.35   & -0.32                 \\
\bottomrule
\end{tabular}
\vspace{-.4cm}
\end{table}

\begin{wrapfigure}{r}{0.5\textwidth}
    \vspace{-.4cm}
    \centering
    \includegraphics[width=\linewidth]{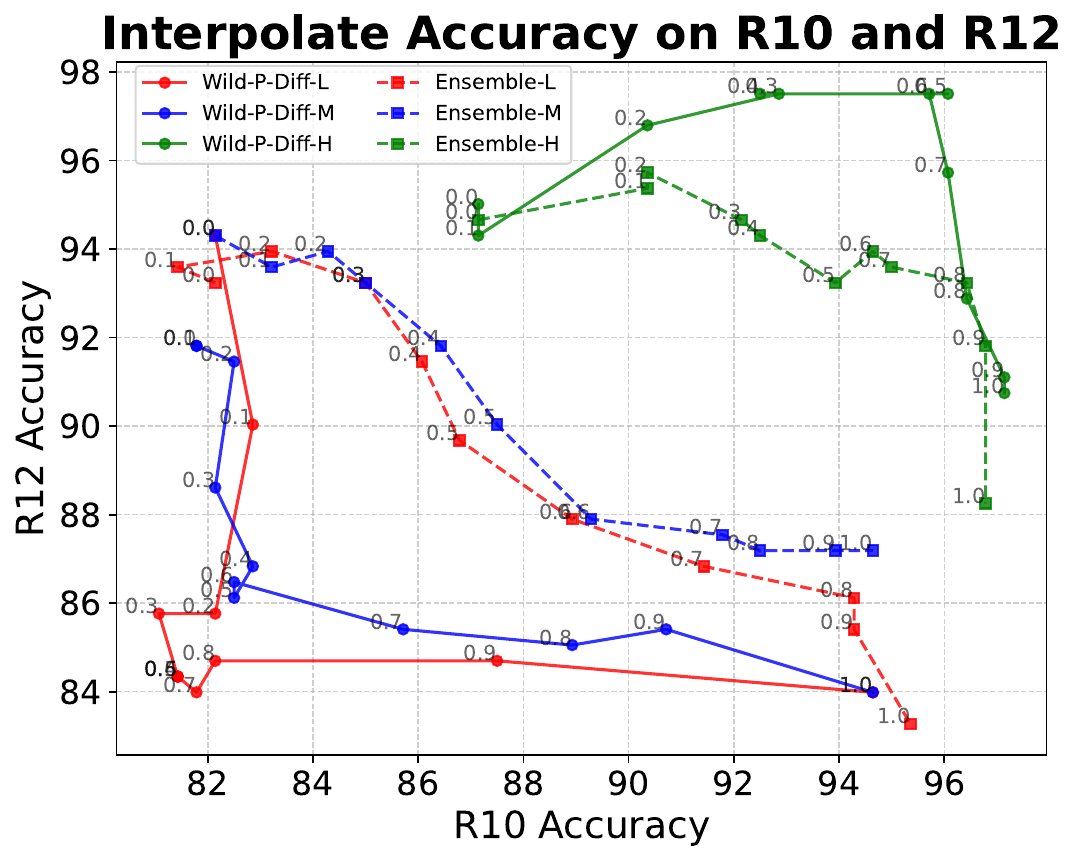}
    \caption{The accuracy of generated parameters when blending two locations' conditions at various interpolation weights (i.e., the solid lines), and that of a baseline method that naively fuses the corresponding fine-tuned parameters of each location called \textit{Ensemble} (i.e., the dashed lines). The values next to the line represent the interpolation weight.
    }
    \label{fig:interpolate}
    \vspace{-.4cm}
\end{wrapfigure}

These results also suggest that generative parameter inference can serve as a compact alternative to storing large numbers of task-specific parameter sets. For example, with 1,000 unique tasks, storing separate LoRA parameters for each (0.34M parameters per task) would require approximately 1.33GB (0.34M × 4 bytes × 1,000). In contrast, a single trained diffusion model with 282.42M parameters requires around 1.10GB—yielding a 15\% memory saving while enabling flexible, on-demand generation.
While \pjn{} relies on task-specific fine-tuning during training to construct the parameter dataset, it eliminates the need to store these parameters explicitly at deployment time. Instead, a single generative model enables efficient retrieval of task-adaptive parameters on demand ($\sim0.81$ seconds per generation pass). This highlights diffusion-based parameter generation as a promising solution for compact and flexible task specialization at scale.

\textbf{RQ2: Inter-Task Interpolation.} 
We next evaluate whether \pjn{} can generate parameters that perform well across multiple tasks by interpolating between task conditions. Specifically, we blend the background features of two known locations at different ratios and use the resulting interpolation as the conditioning input to the diffusion model. The goal is to synthesize parameters that generalize to both locations simultaneously.

Figure~\ref{fig:interpolate} shows the classification accuracy of the generated parameters (\emph{solid} lines) under different interpolation weights between two location conditions. For comparison, the \emph{dashed} lines indicate the performance of a naive baseline (\textit{Ensemble}) that directly fuses the corresponding fine-tuned parameters from each location. We visualize two representative location pairs under all three \pjn{} variants; additional examples are provided in the Appendix.

\begin{wraptable}{r}{0.45\textwidth}
\vspace{-.4cm}
\centering
\scriptsize
\tabcolsep=0.06cm
\caption{The mean and standard deviation of location-specific accuracy on five unseen locations for the three \pjn{} variants and the pre-trained model.} \label{tab:cond-unseen}
\begin{tabular}{c|cccc}
\toprule
\multirow{2}{*}{ID} & \multicolumn{4}{c}{Accuracy}                                                          \\ \cmidrule{2-5} 
                             & \multicolumn{1}{c|}{Pretrain} & \pjn{}-L  & \pjn{}-M  & \pjn{}-H  \\ \midrule
D03                          & \multicolumn{1}{c|}{96.63}     & 96.64 $\pm$ 0.05 & 96.36 $\pm$ 0.22 & 95.62 $\pm$ 0.46 \\
D04                          & \multicolumn{1}{c|}{82.61}     & 82.49 $\pm$ 0.09 & 82.57 $\pm$ 0.26 & 81.43 $\pm$ 0.42 \\
E03                          & \multicolumn{1}{c|}{70.18}     & 70.18 $\pm$ 0.08 & 69.49 $\pm$ 0.48 & 70.13 $\pm$ 0.49 \\
E01                          & \multicolumn{1}{c|}{97.90}     & 97.74 $\pm$ 0.04 & 97.13 $\pm$ 0.31 & 96.21 $\pm$ 0.48 \\
F05                          & \multicolumn{1}{c|}{85.81}     & 86.01 $\pm$ 0.08 & 86.13 $\pm$ 0.27 & 83.05 $\pm$ 0.33 \\
\bottomrule
\end{tabular}
\vspace{-.4cm}
\end{wraptable}

Our results show that \pjn{}-L and \pjn{}-M fail to generate parameters that perform well across both locations, as indicated by the concave shape of their solid curves. 
In contrast, \pjn{}-H, where task-specific parameters exhibit high cross-location similarity demonstrates moderate success. 
For instance, when interpolating between R10 and R12, the green solid line shows that the generated parameters achieve high accuracy on both locations, even outperforming the \textit{Ensemble} baseline.
These findings suggest that \textbf{when task-specific parameters in the training dataset form a coherent and aligned subspace, the diffusion model is able to leverage that structure to interpolate effectively}. In such cases, the model does not merely memorize discrete solutions but learns to sample from a well-behaved region of the parameter space.

\begin{wrapfigure}{r}{0.45\textwidth}
    \vspace{-.4cm}
    \centering
    \includegraphics[width=\linewidth]{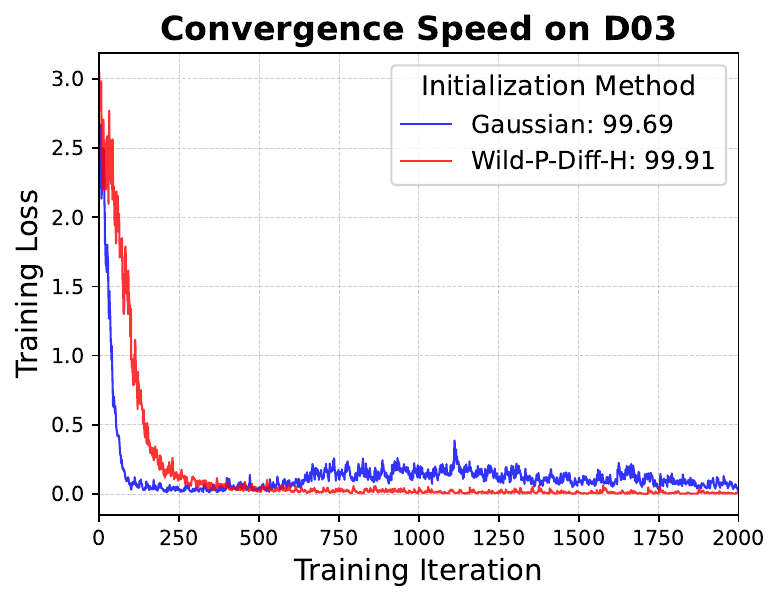}
    \caption{Convergence speed with different LoRA initialization, Gaussian and \pjn{}-H, on unseen location D03. The accuracy after fine-tuning is reported in the legend.}
    \label{fig:convergence}
    \vspace{-.4cm}
\end{wrapfigure}

\textbf{RQ3: Unseen Task Generalization.}
Table~\ref{tab:cond-unseen} reports the validation accuracy on five unseen locations, for which no location-specific parameters are available during training. 
We observe that \textit{none of the \pjn{} variants are able to generate parameters that outperform the pre-trained baseline}, suggesting limited generalization to out-of-distribution tasks. 
The generated LoRA weights yield performance comparable to the pre-trained model, functioning more as an initialization than as a performance gain.

To assess whether these generated parameters can still offer value after fine-tuning, we compare them against a standard Gaussian-initialized LoRA baseline~\cite{hayou2024impact}. 
Figure~\ref{fig:convergence} presents the training loss curves and final classification accuracy for an unseen location D03 (more examples see Appendix Figure~\ref{fig:convergence2}). While both initialization strategies lead to rapid convergence, \pjn{}-H achieves slightly lower final loss and higher final accuracy, indicating a modest improvement.
These results suggest that although diffusion-based parameter generation is not effective for direct deployment on unseen tasks in the current format, it may serve as a useful initialization strategy for downstream fine-tuning, offering mild improvements over traditional random initialization.

\section{Conclusion}
This work explores the use of diffusion models for task-specific parameter generation, where model weights are treated as a new generative modality. 
By conditioning on task identity, the model can reliably generate high-performing parameters for individual tasks seen during training, and moderately support parameter synthesis across multiple tasks via interpolated conditions when the training parameters form a well-aligned subspace.
However, the model fails to generalize to unseen tasks, highlighting a key limitation in its ability to capture out-of-distribution parameter distributions. As such, while this framework reduces deployment-time memory costs and enables on-demand adaptation for known tasks, it does not yet achieve true data-free task adaptation.

\small
\bibliography{reference}
\bibliographystyle{plain}

\newpage
\appendix
\section{Diffusion Model Preliminary}\label{app:diffusion}
Diffusion models~\cite{sohl2015deep}, which our parameter space exploration approach build on, are a class of generative AI models that generate high-resolution images. 
Here we introduce the necessary background about the diffusion model.

\textbf{Forward Diffusion Process.} 
To explain diffusion models, we first explain the forward diffusion process, in which a data point sampled from a real data distribution $x_0 \sim q(x)$ is gradually converted into a noisy representation $x_T$ through $T$ steps of progressive Gaussian noise addition. 
This transformation yields $x_T$ as an isotropic Gaussian noise, i.e., $x_T \sim \mathcal{N}(\mathbf{0},\mathbf{I})$. 
Specifically, the transformation follows the Markov Chain,
\begin{equation}
q(x_t|x_{t-1}) = \mathcal{N}(x_t; \sqrt{1-\beta_t}x_{t-1}, \beta_t \mathbf{I}),
\end{equation}
where $\beta_t \in (0,1)$ is the scheduled noise variance that controls the step size. Here, $\mathcal{N}(x; \mu, \Sigma)$ denotes the probability density function of a Gaussian distribution at the random variable $x$, with mean $\mu$ and covariance matrix $\Sigma$. Therefore, the above transformation states that the conditional distribution of $x_t$ given $x_{t-1}$ is Gaussian, with mean $\sqrt{1 - \beta_t} x_{t-1}$ and covariance matrix $\beta_t \mathbf{I}$.

Then a direct generation of $x_t$ from $x_0$ is:
\begin{equation}\label{eq:forward-diffusion-closed}
x_t = \sqrt{\bar{\alpha}_t}x_0 + \sqrt{1-\bar{\alpha}_t}\epsilon,
\end{equation}
where $\bar{\alpha}_t=\prod^t_{i=0}(1-\beta_i)$, $\beta_i \in (0,1)$ is the scheduled noise variance that controls the step size, and $\epsilon \sim \mathcal{N}(0,\mathbf{I})$.

\textbf{Reverse Process.} 
Diffusion models reverse this forward process, learning to retrieve the original image $x_0$ from the noise $x_T$ by estimating the noise at each step and iteratively performing denoising.
The Denoising Diffusion Implicit Model (DDIM) \cite{song2020denoising} is a prominent denoising method, known for its efficiency and deterministic output. 
It requires fewer steps, sometimes only 50, to replicate the denoising achieved by the standard 1000-step process, and consistently reproduces $x_0$ from a given $x_T$, providing deterministic reconstruction.
Formally, for each denoising step $t$, a learned noise predictor $\epsilon_\theta(\cdot)$ estimates the noise $\epsilon$ added to $x_0$, leading to an approximation of $x_0$.
\begin{equation}
    x'_0 = \frac{x_t - \sqrt{1-\bar{\alpha}_t} \epsilon_\theta(x_t, t)}{\sqrt{\bar{\alpha}_t}}.
\end{equation}
Then DDIM reintroduces $\epsilon_\theta(x_t, t)$ to determine $x_{t-1}$: 
\begin{equation}\label{eq:diffusion_xt_xt-1}
    x_{t-1}=\sqrt{\bar{\alpha}_{t-1}} (x'_0)+\sqrt{1-\bar{\alpha}_{t-1}}\epsilon_\theta(x_t, t).
\end{equation}
In this way, DDIM could deterministically recover the same image $x_0$ from the specified noise $x_T$. 

\textbf{Diffusion Model Training and Inference.}
Diffusion models approximate the reverse diffusion process through a neural network $\epsilon_\theta(\cdot)$, which is trained to estimate the noise component at each denoising step. The model takes a noisy input $x_t$ and its timestep $t$ to predict the noise added at that step.
During training, the noisy sample is generated using Equation~\ref{eq:forward-diffusion-closed} from the clean training dataset. The model is then trained with the following objective:
\begin{equation}\label{eq:diffusion-train}
\mathcal{L}(\theta)=\mathbb{E}_{t\sim[1,T],\epsilon \sim \mathcal{N}(0,\mathbf{I})} ||\epsilon - \epsilon_\theta(x_t,t)||_2^2,
\end{equation}
where $t$ is sampled uniformly from 1 to $T$. The typical architecture of the noise estimator is a UNet \cite{ronneberger2015unet}. 

During inference, we first sample a random Gaussian noise $x_T \sim \mathcal{N}(\mathbf{0},\mathbf{I})$ and then iteratively compute $x_{T-1}$ using Equation~\ref{eq:diffusion_xt_xt-1} until the clean output $x_0$ is obtained.
For stable diffusion, the key difference lies in operating within the latent space rather than the image space. Specifically, all $x$ variables are replaced by their latent representations $\mathbf{Z}$, with a pre-trained Variational Autoencoder (VAE) performing the transformation between the original space and the latent space.

\section{Additional Experimental Settings}

\subsection{Data Statistics}
\begin{table}[htb]
    \centering
    \caption{Serengeti Safari Camera Trap dataset with train and test split.}\label{tab:dataset_stats}
    \begin{tabular}{cc|cc}
        \toprule
        \multicolumn{2}{c|}{\textbf{Train}} & \multicolumn{2}{c}{\textbf{Test}} \\
        \# Locations & \# Images & \# Locations & \# Images \\
        \midrule
        110 & 100,289 & 5 & 15,051 \\
        \bottomrule
    \end{tabular}
\end{table}

\begin{table}[htb]
\centering
\caption{Fine-tuning locations and their data statistics. The locations are sorted with the number of animal images (i.e., \#images w/o empty).} \label{tab:finetune-locs}
\begin{tabular}{c|cc|cc}
\toprule
\multirow{2}{*}{Location ID} & \multicolumn{2}{c|}{\#images} & \multirow{2}{*}{\#train} & \multirow{2}{*}{\#eval} \\
                             & w/ empty     & w/o empty     &                          &                         \\ 
\midrule
R10                          & 2795         & 1766          & 2515                     & 280                     \\
R12                          & 2803         & 1683          & 2522                     & 281                     \\
U11                          & 306          & 210           & 275                      & 31                      \\
Q08                          & 291          & 201           & 261                      & 30                      \\
M08                          & 195          & 156           & 175                      & 20                      \\
D09                          & 228          & 153           & 205                      & 23                      \\
H13                          & 171          & 96            & 153                      & 18                      \\
S09                          & 147          & 57            & 132                      & 15                      \\
U13                          & 403          & 51            & 362                      & 41                      \\
S12                          & 208          & 51            & 187                      & 21                      \\ 
\bottomrule
\end{tabular}
\end{table}

\subsection{Model Architecture Details}\label{app:architecture}
\textbf{Parameter Encoding VAE.}
In practice, the encoder $\mathcal{E}$ is a 1D CNN-based architecture with channel configurations of (64, 128, 256, 256, 32). At the final layer, the features are flattened and projected to a latent dimension of 2048 via a linear transformation.
The decoder $\mathcal{D}$ employs transposed convolutions with the same channel configuration to reconstruct the original parameter vector.

\textbf{Parameter Generation UNet.}
The architecture follows a standard encoder-decoder structure, consisting of multiple 1D convolutional layers with batch normalization (BN). The channel configuration is set as (1, 64, 128, 256, 512, 256, 128, 64, 1), ensuring progressive downsampling and upsampling within the latent space. 
Additionally, the timestep $t$ is encoded using positional encoding \cite{vaswani2017attention} and directly added to the latent representation $\mathbf{Z}_t$ to guide the model in predicting noise distributions across different timesteps.

\begin{table}[htb]
\centering
\caption{Training recipe for parameter autoencoder and diffusion model.} \label{tab:training_recipes}
\begin{tabular}{l|cc}
    \toprule
    Config & \begin{tabular}[c]{@{}l@{}}Parameter\\ Autoencoder\end{tabular} & \begin{tabular}[c]{@{}l@{}}Diffusion\\ UNet\end{tabular} \\
    \midrule
    Optimizer & \multicolumn{2}{c}{AdamW} \\
    Learning rate (LR) & $1e^{-5}$ & $1e^{-4}$ \\
    Weight decay & \multicolumn{2}{c}{$1e^{-5}$} \\
    Training iterations & 20000 & 30000 \\
    Batch size & 16 & 50 \\
    LR schedule & \multicolumn{2}{c}{Cosine decay} \\
    \bottomrule
\end{tabular}
\end{table}

\subsection{The Selection of Task-Specific Parameters}\label{sect:what-param}
We empirically identify which subset of parameters should be treated as task-specific to achieve accuracy comparable to fine-tuning the entire pre-trained model. 

\begin{table}[htb]
\centering
\caption{Impact of location-specific parameter selections for location \textit{R10}.
Incorporating LoRA adapters in the first 6 layers (i.e., \textit{First 6 layers LoRA}) achieves a validation accuracy comparable to fully fine-tuning all parameters (i.e., \textit{ALL}) with only 8.8\% trainable parameters (i.e., $0.34/3.85$).} \label{tab:R10-trials}
\begin{tabular}{ccc}
\toprule
Fine-tuned Param.            & \#Params. (M) & Accuracy        \\
\midrule
ALL                          & 3.85          & \textbf{0.9571} \\
BN                           & 0.04          & 0.8107          \\
First 5 layers               & 0.29          & 0.8893          \\
First 6 layers               & 0.81          & 0.9179          \\
First 7 layers               & 2.74          & 0.9500          \\
Last 1 layer                 & 0.39          & 0.8071          \\
Last 2 layers                & 1.08          & 0.8393          \\ 
Last 3 layers                & 3.01          & 0.9321          \\
\textbf{First 6 layers LoRA} & \textbf{0.34} & \textbf{0.9500}   \\
\bottomrule
\end{tabular}
\end{table}

Table~\ref{tab:R10-trials} reports the validation accuracy and parameter counts for location \textit{R10}, exploring different strategies for selecting location-specific parameters. 
We consider (1) assigning several initial or final layers as location-specific, (2) designating only the Batch Normalization layers~\cite{mudrakarta2018k}, and (3) introducing LoRA adapters~\cite{hu2022lora}. 
We observe that selecting location-specific parameters in the initial layers yields superior accuracy compared to using the last two or three layers, despite involving fewer parameters.

Most importantly, we find that the optimal approach is to insert location-specific LoRA adapters in the first six layers, obtaining an accuracy of 95\%, nearly identical to fine-tuning all parameters, while requiring only 8.8\% number of the entire parameters. 
Table~\ref{tab:loc-lora} further confirms the high accuracy achieved for different locations.
Therefore, in the following experiments, we focus on generating these identified LoRA parameters with diffusion models for each location.

\begin{table}[htb]
\centering
\caption{Validation accuracy with introducing location-specific LoRA adapters in the first six layers for different locations, comparing with fine-tuning all parameters (i.e., \textit{ALL}).} \label{tab:loc-lora}
\begin{tabular}{c|ccc}
\toprule
\multirow{2}{*}{Location ID} & \multicolumn{3}{c}{Accuracy}            \\
                             & Pretrain & ALL  & First 6 layers LoRA \\
\midrule
R10                          & 0.81       & 0.96 & 0.95                \\
R12                          & 0.85       & 0.96 & 0.95                \\
U11                          & 0.45       & 1.00 & 1.00                \\
Q08                          & 0.57       & 0.97 & 0.90                \\
M08                          & 0.65       & 1.00 & 1.00                \\
D09                          & 0.87       & 1.00 & 1.00                \\
H13                          & 0.78       & 0.94 & 0.94                \\
S09                          & 1.00       & 1.00 & 1.00                \\
U13                          & 0.17       & 1.00 & 1.00                \\
S12                          & 0.95       & 1.00 & 1.00                \\
\bottomrule
\end{tabular}
\end{table}

\section{Unconditional Parameter Generation Cont.}\label{sect:exp-uncond-param-cont}
\textbf{Fine-Tuned Models Diversity.}
Recall that after each location-specific model converges, we continue fine-tuning for additional iterations and save 300 checkpoints. Let $M_0$ denote this initial, converged model before additional fine-tuning begins. 
A \emph{saving interval} of 1 stores a checkpoint at every iteration (300 total), while intervals of 10 or 100 accumulate 3,000 or 30,000 extra iterations, respectively. 
Intuitively, larger saving intervals provide more substantial perturbations in the parameters between consecutive checkpoints, thus increasing the fine-tuned model diversity in the train set.

\begin{figure}[htb]
  \begin{subfigure}[t]{0.45\linewidth}
    \centering
    \includegraphics[width=1\linewidth]{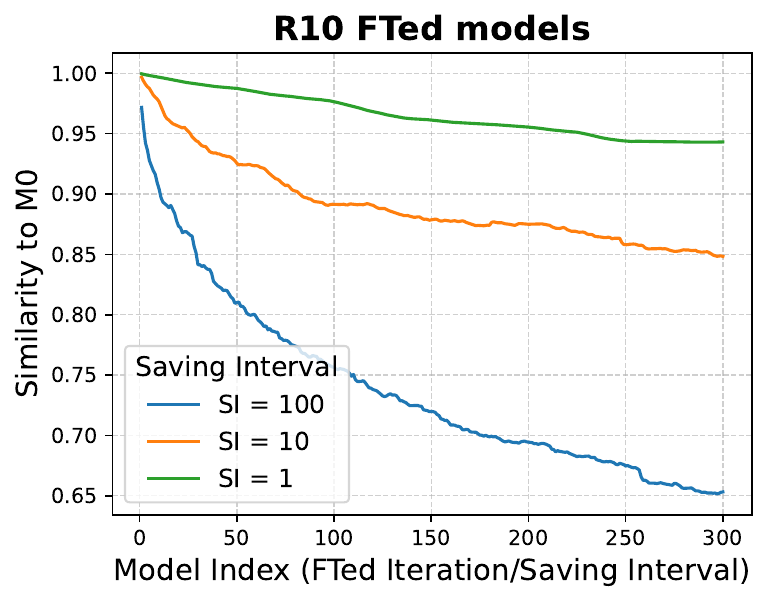}
    \caption{Similarity between each fine-tuned model and $M_0$, along with the model index.}
  \end{subfigure}
  \hfill
  \begin{subfigure}[t]{0.52\linewidth}
    \centering
    \includegraphics[width=1\linewidth]{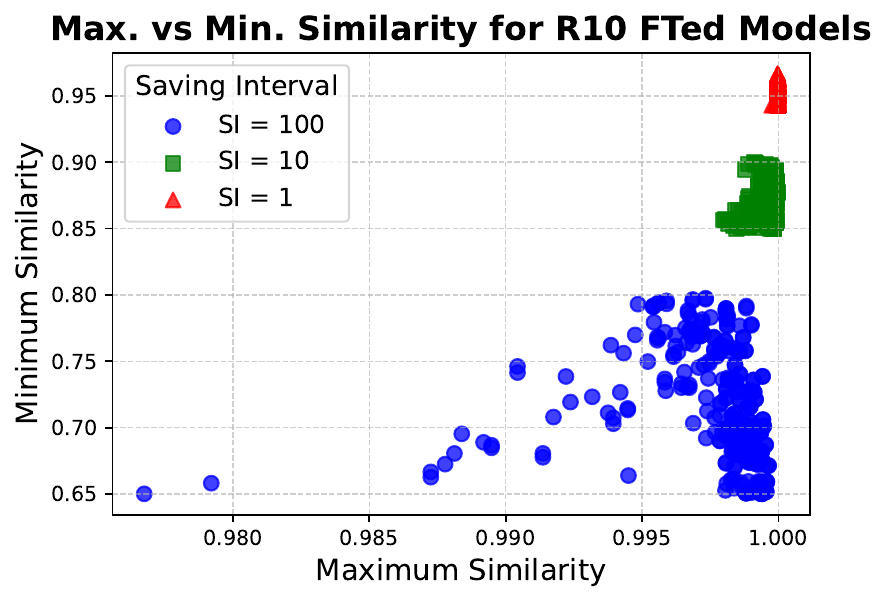}
    \caption{Maximum and minimum similarity between 300 fine-tuned checkpoints.}
  \end{subfigure}
  \caption{Input diversity for location \textit{R10} for different saving intervals, i.e., \textit{SI = 1}, \textit{SI = 10}, \textit{SI=100}.}
  \label{fig:uncond-input-diversity}
\end{figure}

Figure~\ref{fig:uncond-input-diversity} illustrates how these varying saving intervals influence the diversity of the 300 fine-tuned checkpoints for location \emph{R10}. 
In Figure~\ref{fig:uncond-input-diversity}(a), we plot the similarity between each checkpoint and $M_0$, revealing that larger saving intervals lead to more deviation from the original converged model, boosting overall diversity. 
Correspondingly, Figure~\ref{fig:uncond-input-diversity}(b) shows the maximum and minimum similarity among those 300 checkpoints. 
Here, for each checkpoint, we compute the cosine similarity to all other checkpoints and record its maximum and minimum values.
This provides a measure of how close or distant each checkpoint is relative to the rest, reflecting the overall spread and diversity of the parameter set.
The results confirm that checkpoints derived using higher saving intervals exhibit more spread, i.e., higher diversity.

Having clarified how saving intervals modulate the variety of fine-tuned checkpoints, we now explore whether \pjn{} simply memorizes these reference models or truly generates novel parameters under different configurations.

\textbf{Generated Models Novelty and Diversity.}
To address whether \pjn{} merely reproduces existing fine-tuned checkpoints or creates genuinely new parameters, we measure both \emph{output novelty} and \emph{output diversity} for the generated models. 
Specifically, as introduced in Section~\ref{sect:pdiff-exp-setting}, we compute (1) the maximum similarity of each generated parameter to every fine-tuned checkpoint, and (2) the maximum similarity among the generated parameters themselves.

\begin{figure}[htb]
    \centering
    \includegraphics[width=0.8\textwidth]{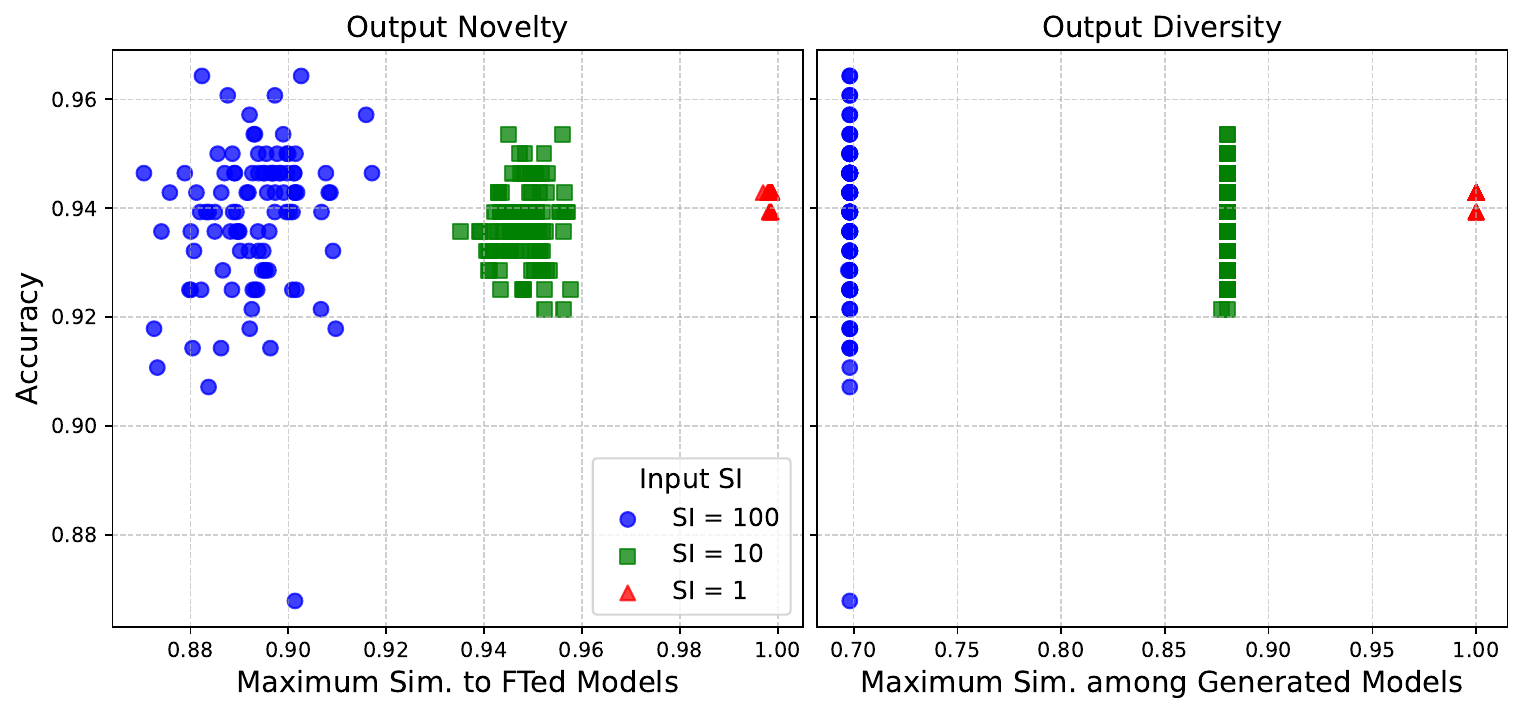}
    \caption{The output novelty and diversity are influenced by the different saving intervals used when collecting the training checkpoints.}
    \label{fig:output_novelty_diversity}
\end{figure}

In Figure~\ref{fig:output_novelty_diversity}, our results reveal a clear distinction based on the input diversity. 
When the saving interval is small (the red points), i.e., the input diversity is low, the generated parameters exhibit high similarity (almost 1) to the fine-tuned checkpoints, which suggests memorization, and also to one another, which indicates limited variety. 
However, when the saving interval is larger (the blue points), namely the input diversity is higher, the generated parameters have lower similarity to the fine-tuned checkpoints for about 0.89. 
This gap implies that \pjn{} is sampling genuinely new samples from the learned parameter distribution rather than simply replicating existing models. 
In this case, the output diversity also increases, as reflected by lower similarity among generated parameters.

Overall, these findings show that \textbf{high diversity among fine-tuned checkpoints facilitates novel parameter generation in \pjn{}}. By varying the saving interval, we can tune the input model diversity, which in turn influences whether the diffusion model memorizes or creatively explores the parameter distribution.

\section{Conditional Parameter Generation Cont.}\label{sect:exp-cond-param-cont}
\textbf{Parameter Similarity across Locations}
We vary the initial converged model, denoted $M_0$, from which fine-tuning checkpoints are derived for each location, thereby inducing different cross-location similarities:
\begin{itemize}[noitemsep,nolistsep,topsep=0pt]
    \item \pjn{}-L: Each location has an \emph{independent} $M_0$, meaning each LoRA component is initialized randomly for each location. 
    \item \pjn{}-M: Locations share a \emph{similar} $M_0$, with all LoRA parameters initialized identically across locations before location-specific fine-tuning.
    \item \pjn{}-H: Locations use the \emph{same} $M_0$ precisely, i.e., the LoRA parameters are first fine-tuned on one particular location and then used as a starting point for all the others when saving multiple checkpoints per location.
\end{itemize}

\begin{figure}[htb]
    \centering
    \includegraphics[width=0.32\textwidth]{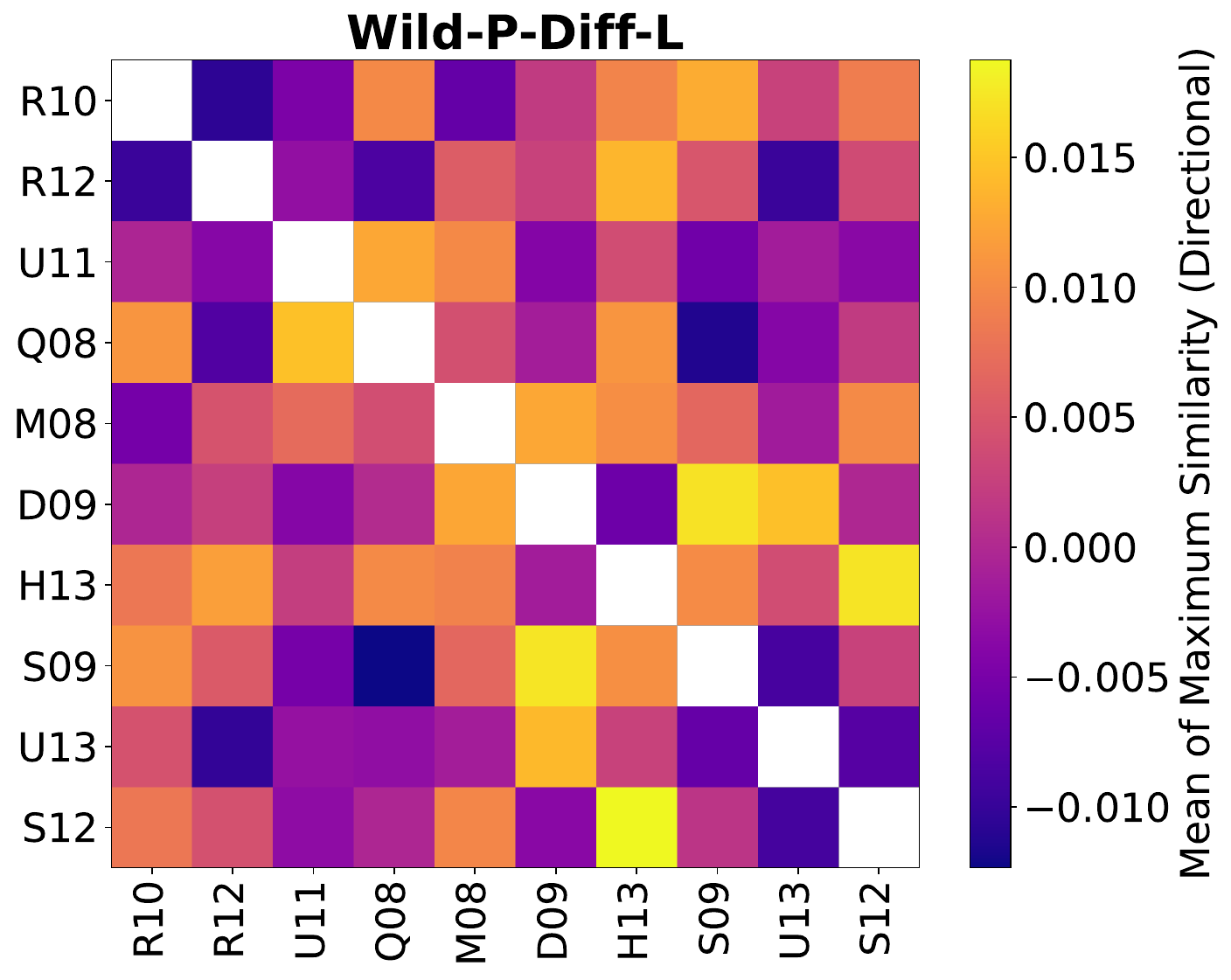} 
    \includegraphics[width=0.3\textwidth]{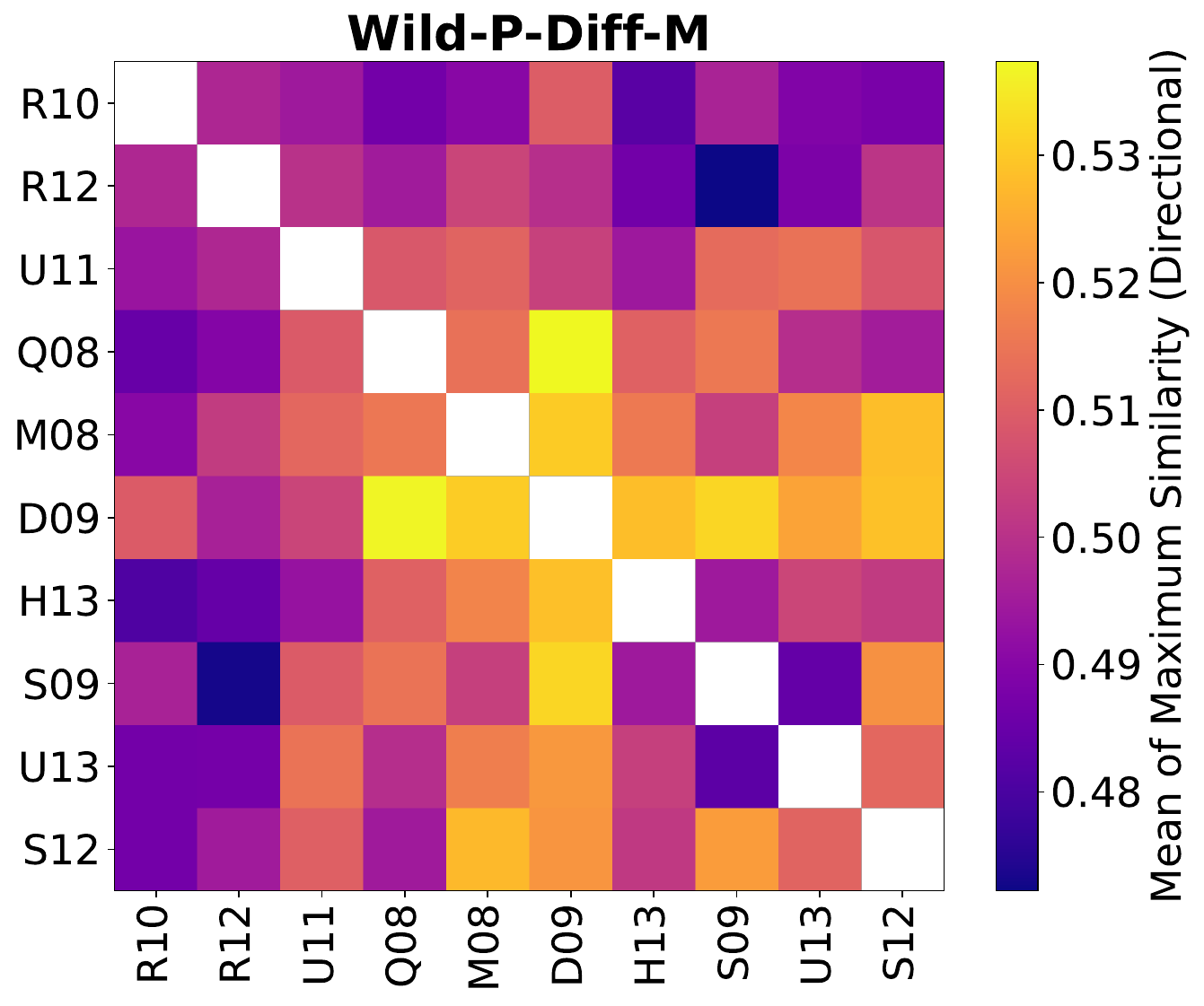}
    \includegraphics[width=0.3\textwidth]{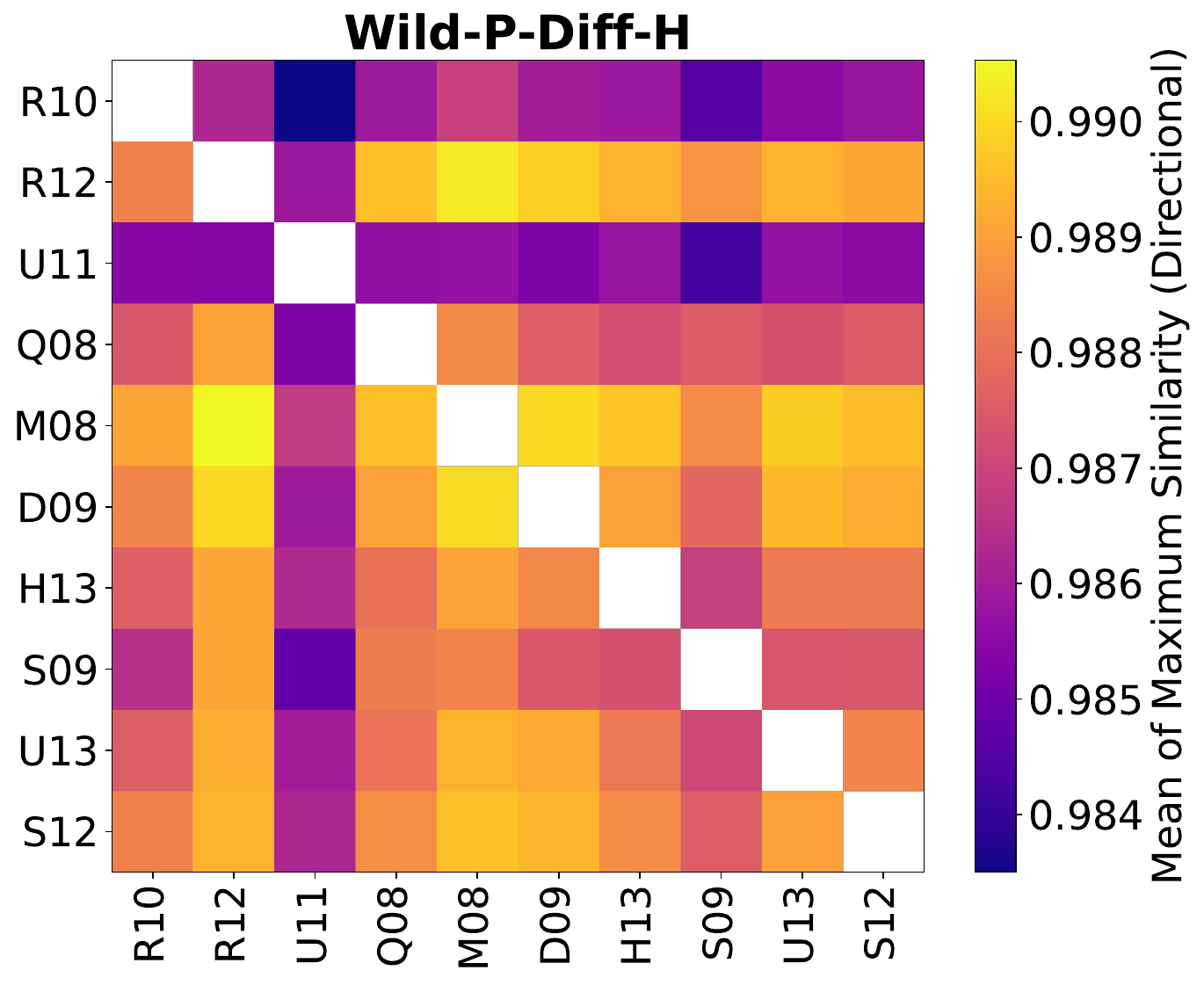}
    \caption{Pairwise similarity of input checkpoints across different locations. \pjn{}-L, \pjn{}-M, and \pjn{}-H correspond to the three cross-location similarity strategies used when collecting location-specific fine-tuned checkpoints, respectively.}
    \label{fig:IDAL}
\end{figure}

By modifying how $M_0$ is set for different locations, we obtain three different levels of cross-location similarity. 
Specifically, for any pair of locations \textit{loc1} and \textit{loc2}, we compute the maximum similarity across their respective fine-tuned checkpoints and then average these values. 
Figure~\ref{fig:IDAL} summarizes the cross-location similarity under each variant of \pjn{}, where we have three observations.
First, \emph{\pjn{}-L} yields minimal similarity across locations (close to 0), implying almost orthogonal LoRA parameters across locations in the space.
Second, \emph{\pjn{}-M} increases the cross-location similarity to around 0.5.
Last, \emph{\pjn{}-H} raises it further to approximately 0.98, indicating that location-specific parameters are extremely close to each other.

\begin{figure}[htb]
    \centering
    \includegraphics[width=0.48\textwidth]{Figures/interpolate_acc_R10_R12.pdf}
    \includegraphics[width=0.48\textwidth]{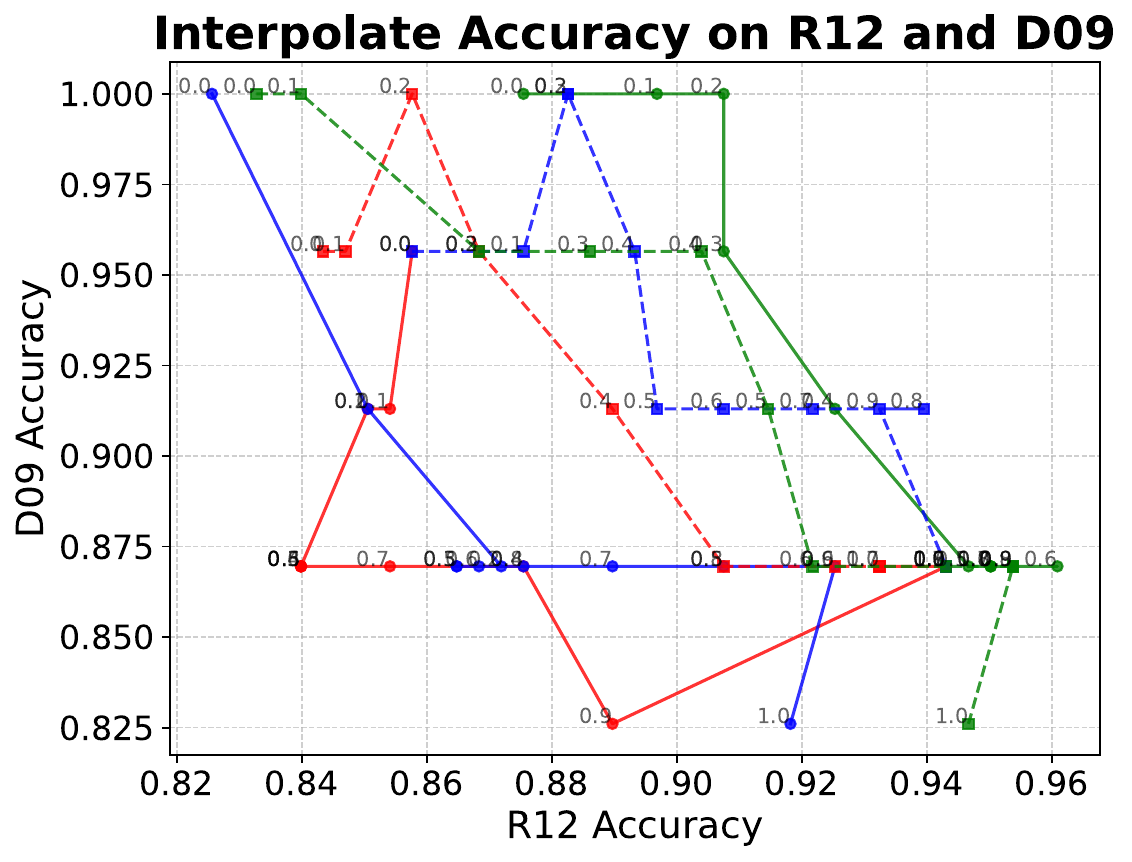} 
    \includegraphics[width=0.48\textwidth]{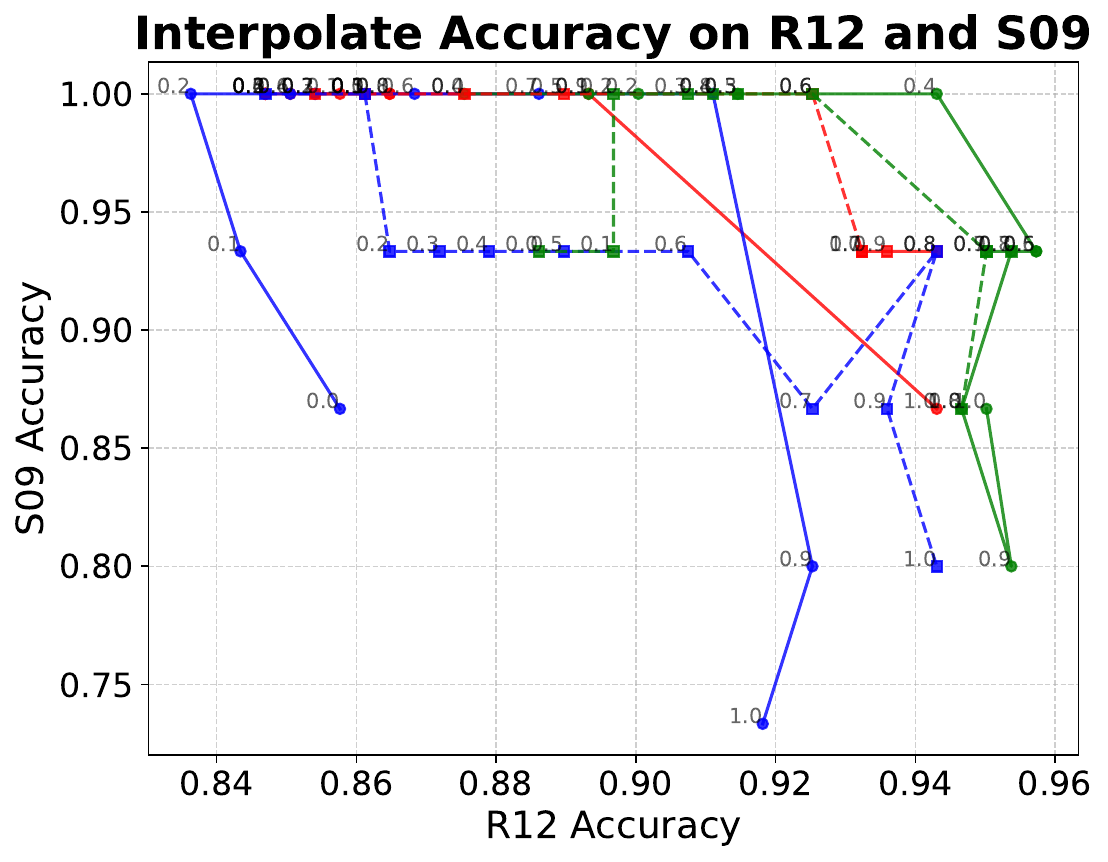}
    \includegraphics[width=0.48\textwidth]{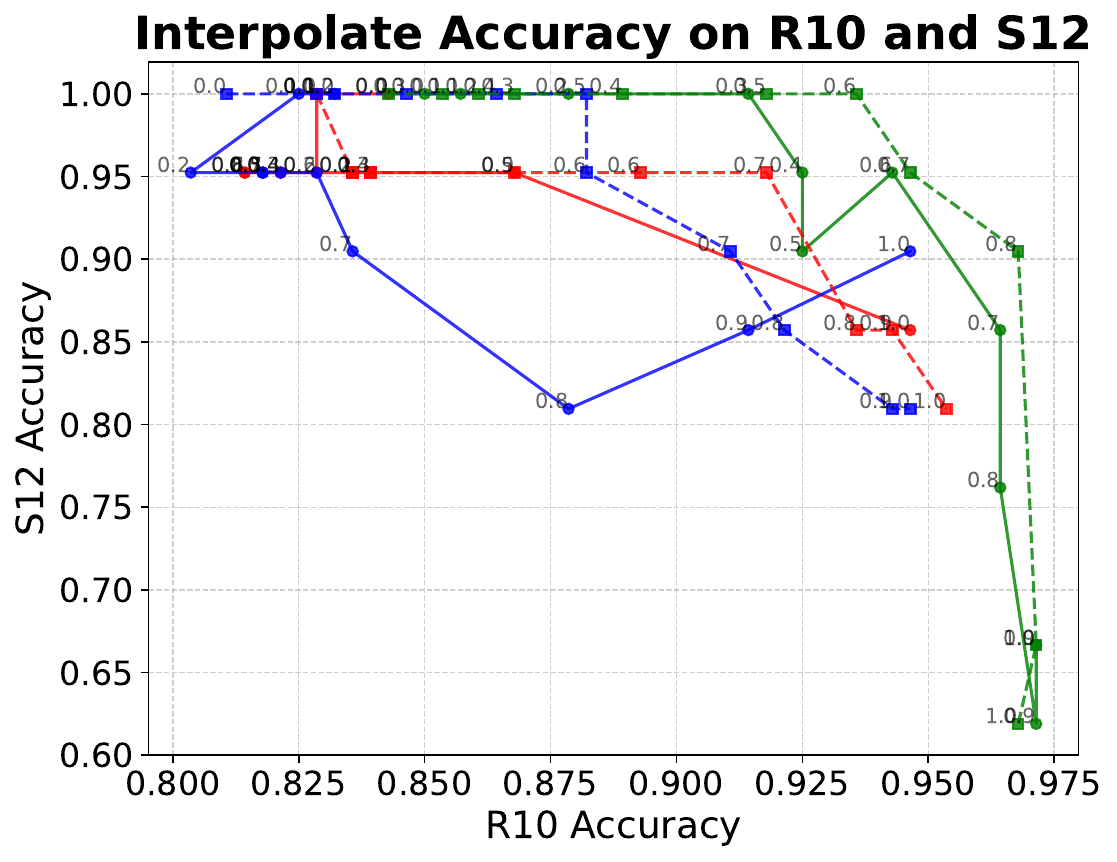}
    \caption{More examples for Inter-Location Interpolation.} 
    \label{fig:interpolate2}
\end{figure}

\textbf{RQ2: Inter-Task Interpolation.}
Figure~\ref{fig:interpolate2} exhibits more examples for generating parameters that work for multiple tasks. As mentioned in the main paper, by interpolating the condition of two locations, \pjn{}-H could successfully interpolate the parameter space to generate parameters for multi-task use.

\begin{figure}[htb]
    \centering
    \includegraphics[width=0.48\textwidth]{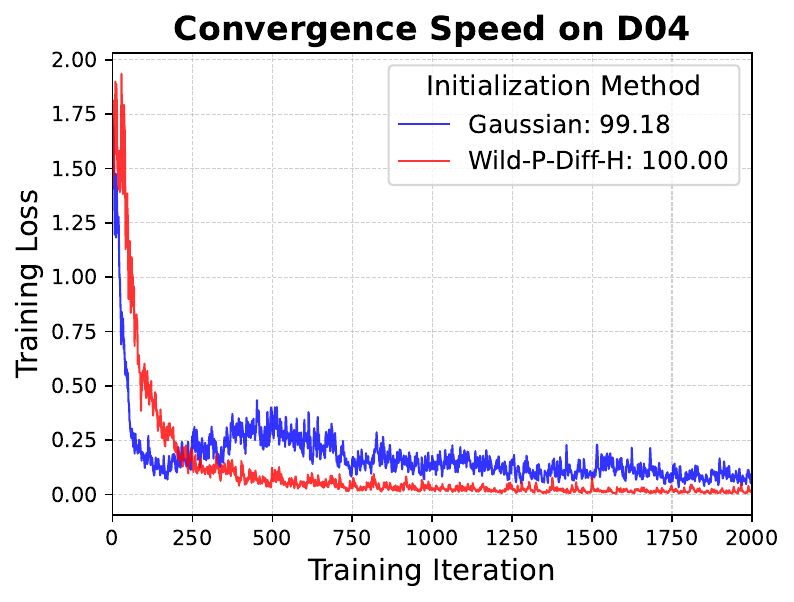}
    \includegraphics[width=0.48\textwidth]{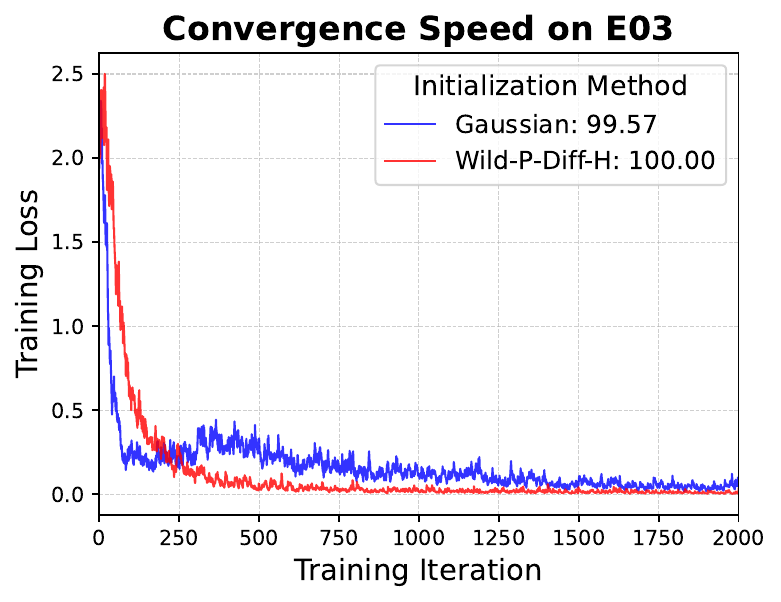}
    \caption{Convergence speed with different LoRA initialization, Gaussian and \pjn{}-H, on two additional unseen locations, D04 and E03. The accuracy after fine-tuning is reported in the legend.}
    \label{fig:convergence2}
\end{figure}

\textbf{RQ3: Unseen Task Generalization.}
We provide more examples when using diffusion-generated parameters as the LoRA initialization weights for two additional unseen locations, D04 and E03, in Figure~\ref{fig:convergence2}. As mentioned in the main text, we observed that \pjn{}-H leads to slightly lower final training loss and higher validation accuracy compared to the conventional Gaussian initialization, indicating the potential of using diffusion-generated parameters for unseen tasks as a new initialization strategy.

\end{document}